\documentclass{article}
\usepackage{amsfonts}
\usepackage{amsmath,graphicx,mlspconf}
\usepackage{todonotes} 
\usepackage{enumitem}
\newlist{deflist}{description}{2}
\setlist[deflist]{labelwidth=2cm,leftmargin=!,font=\normalfont}

\def\textsc#1{\textnormal{{\sc #1}}}%

\copyrightnotice{979-8-3503-2411-2/23/\$31.00 {\copyright}2023 IEEE}

\toappear{2023 IEEE International Workshop on Machine Learning for Signal Processing, Sept.\ 17--20, 2023, Rome, Italy}

\title{Low-count time series anomaly detection}
\twoauthors{%
  Philipp Renz, \sthanks{Work conducted during employment at Amazon.}
}{%
    Johannes Kepler University, Linz
}{%
  Kurt Cutajar, Niall Twomey, Gavin K.\ C.\ Cheung and Hanting Xie
}{%
    Amazon Prime Video, UK
}

\name{%
    Philipp Renz$^{{\star},1}$\thanks{${}^1$Work conducted during an internship at Amazon Prime Video, UK.}%
    \qquad Kurt Cutajar$^{\dagger}$%
    \qquad Niall Twomey$^{\dagger}$%
    \qquad Gavin K.\ C.\ Cheung$^{\dagger}$%
    \qquad Hanting Xie$^{\dagger}$%
}
\address{%
    $^{\star}$ Johannes Kepler University Linz, Austria\qquad$^{\dagger}$ Amazon Prime Video, UK%
}

\begin{document}

\maketitle

\begin{abstract}
Low-count time series describe sparse or intermittent events, which are prevalent in large-scale online platforms that capture and monitor diverse data types.
Several distinct challenges surface when modelling low-count time series, particularly low signal-to-noise ratios (when anomaly signatures are provably undetectable), and non-uniform performance (when average metrics are not representative of local behaviour).
The time series anomaly detection community currently lacks explicit tooling and processes to model and reliably detect anomalies in these settings.
We address this gap by introducing a novel generative procedure for creating benchmark datasets comprising of low-count time series with anomalous segments.
Via a mixture of theoretical and empirical analysis, our work explains how widely-used algorithms struggle with the distribution overlap between normal and anomalous segments.
In order to mitigate this shortcoming, we then leverage our findings to demonstrate how anomaly score smoothing consistently improves performance.
The practical utility of our analysis and recommendation is validated on a real-world dataset containing sales data for retail stores.
\end{abstract}
\section{Introduction}
\label{sec:intro}

\begin{figure}[t!]
  \centering
  \centerline{\includegraphics[width=8.5cm]{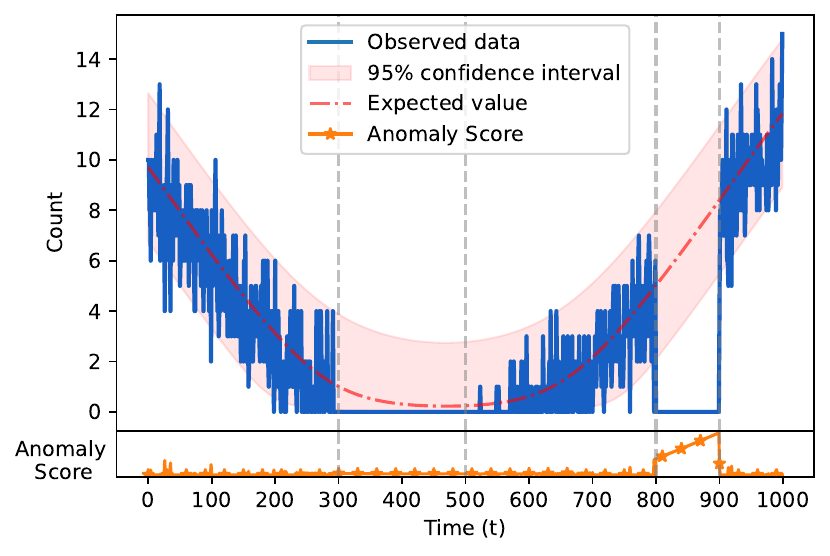}}
  \vspace{-0.4cm}
\caption{Example of a forecasting-based anomaly detection model applied to a low-count time series. Two drop-to-zero anomalies are injected in the intervals $[300, 500]$ and $[800, 900]$. The anomaly score is non-zero when the observed data is outside the forecasted 95\% confidence interval.}
\label{fig:forecasting_failed}
\vspace{-.5cm}
\end{figure}

Time Series Anomaly Detection (TSAD) is a highly prevalent research area, with success stories in multiple application domains. 
However, solutions developed in one domain do not directly translate to success in others, and as a result TSAD has been explored with broad perspectives. 
One of the most successful applications of large-scale TSAD has been in monitoring e-commerce, where it provides essential health check support for production services. 
Indeed, in the case of web-scale service monitoring, TSAD is often the only recourse to assure service reliability since humans do not have the capacity or ability to monitor millions of fragmented time series. 

As well as the sheer volume of time series to be monitored, the diversity of time series could also preclude the effectiveness of an off-the-shelf TSAD algorithm. 
This diversity can manifest in many forms, including data type, sampling rate, seasonality and count levels. 
Each of these can present at different degrees, and together create an implicit spectrum of data with varying performance characteristics. 
Large datasets may consist of mixtures of multiple long-tail effects, each of which carries unique accuracy profiles. 
However, performance evaluation rarely stratifies datasets across these important traits. 
A specific failure case on a low-count time series is illustrated in Figure~\ref{fig:forecasting_failed}. 
Here, a forecasting-based AD model is unable to distinguish an obvious anomaly at $t\in[300, 500]$ since its signature falls within the expected prediction bands.
Conversely, the anomaly at $t\in[800,900]$ is detectable.
As demonstrated by this simple example, global average performance may lead practitioners to have optimistic performance expectations in these contexts, unless specific attention is given to these edge cases. 

The literature on TSAD is varied and prolific, but the notion of what constitutes an anomaly is often only loosely defined.
As a result, even thorough surveys and benchmarks  \cite{schmidlAnomalyDetectionTime2022,paparrizosTSBAUDEndtoEndAnomaly2022} may not accurately depict how well different methods perform in specific contexts.
Our work in this paper focuses on improving model performance along the long-tail of low-count seasonal time series. 
While low-count time series may be perceived as describing services that are less important or significant, these often constitute a heavy tail in the distribution of time series, resulting in a substantial portion of data having lower monitorability. 
Moreover, collections of time series may exhibit intermittent behaviour at granular time intervals in spite of tracking business-critical services, such as sales of expensive items, user conversion, etc.

To the best of our knowledge, no previous work has specifically paid attention to how TSAD methods perform in low-count time series.
In this paper, we are interested in assessing the resilience of AD algorithms across the full spectrum of high- to low-count metrics, where the overlap between expected and anomalous data distributions increases the complexity of accurate and timely detection.
In particular, (i) we introduce a novel data simulation procedure for seasonal time series having different count levels and anomalous segments.
(ii) We leverage this to construct a benchmark exploring disparities in performance, and (iii) recommend a post-hoc strategy based on anomaly score smoothing which improves performance without compromising time-to-detection.
Finally, (iv) we outline general guidance on how our contributions can be used in new contexts.
\section{Related Work}
\label{sec:related_works}
Several works have focused on evaluating and comparing the performance of TSAD methods \cite{schmidlAnomalyDetectionTime2022, paparrizosTSBAUDEndtoEndAnomaly2022}.
While these provide helpful insights for solving general time series monitoring problems, recommendations from these studies do not specifically address challenges related to low-count time series.

Few studies have proposed specific approaches for anomaly detection in similarly challenging set-ups.
In the work by Shipmon et al.~\cite{shipmonTimeSeriesAnomaly2017}, a forecasting-based method is presented to detect drops in noisy and sparse data.
Similarly, Fan et al.~\cite{fanUnsupervisedAnomalyDetection2023} propose a multi-granular approach for detecting demand changes in intermittent data, which prove effective for their specific problems.
While these studies propose interesting methods for detecting anomalies in low-count or noisy time series, they do not provide a general overview of how different TSAD model classes perform on low-count time series, and nor do they develop concrete understanding on the shortcomings of standard approaches. 
Change point detection~\cite{denBurg2020} may also be leveraged to detect persistent changes in data distribution, but does not explicitly differentiate between normal and anomalous behaviour across the life cycle of a time series, particularly with low signal-to-noise ratio.
\section{A New Benchmark for Low-Count TSAD}

One of the primary reasons for which AD in low-count time series has been underexplored is the lack of an established benchmark dataset covering the required diversity of count-based time series.
In view of this shortcoming, the first contribution of this work is the specification of a generative process for synthesising count-based time series with the flexibility to control key data properties (count,  noise and anomalies).

In the general TSAD setting, we observe a series of values, $x_{1:t} = \left[ x_1, \dots, x_t\right]$, which describe a system of interest at time steps $\{1,\dots, t\}$.
The objective is to generate an anomaly score $a_t(x_{1:t})$ at each time step based on how new data deviate from expected patterns.
Thresholds can then be applied to the anomaly scores for optimal decision making. 

\subsection{Time Series Generation}\label{sec:synthetic_data}

For obtaining normal count time series, we simulate seasonal behaviour via a transformation of a cosine wave:
\begin{equation}\label{eq:cosine}
    \lambda_t = A * \Delta t\left(1 + \cos\left(2 \pi * f * t *\Delta t\right)\right) / \,2,
\end{equation}
\noindent where $A$ is an amplitude parameter that controls the peak of the wave, $f$ denotes the frequency of the seasonal pattern, and $\Delta t$ the time interval at which the data is sampled.
More generally, one could use any periodic function such as a Fourier series to model seasonal behaviour, but for the purposes of this work, we elect to use the cosine function due to its simplicity.
The resulting value at each time step, $\lambda_t$, is then used as the mean of a Poisson distribution in order to sample count values of the time series $X_t \sim\mathcal{P}\left(\lambda_t\right)$.
We can control expected count levels of the time series by adjusting $A$. 

\subsection{Anomaly Injection}\label{sec:anomaly_injection}

Motivated by our interest in longer-running contextual anomalies, the injection of anomalies is achieved by a two-state Markov Chain method similar to that introduced in \cite{alexandrovGluonTSProbabilisticNeural2020}.
We begin by specifying a transition matrix $T \in \mathbb{R}^{2 \times 2}$, where $T_{ij}$ specifies the probability $T_{ij} = P\left(S_t=j|S_{t-1}=i\right)$.
To simulate the real-world expectation that anomalies are typically rare or infrequent, the transitional probabilities are intentionally specified to increase the likelihood of remaining in a normal state rather than transition into an anomalous one.
The emission distribution's rate parameter is adjusted in an abnormal state by a \emph{reduction rate}, $r\in \left[ 0,1\right]$, and the generative model is given by $X_t \sim \mathcal{P}((1-r) \lambda_t)$. Usual behaviour is recovered in non-anomalous states when $r = 0$. 
Samples from our model achieve low signal-to-noise ratios at low rates since the signal-to-noise ratio of a Poisson random variable is proportional to $\sqrt{\lambda}$.
%

\subsection{Experimental Dataset}\label{sec:exp_dataset}

\begin{figure}[t!]
  \centering
  \centerline{\includegraphics[width=8.5cm]{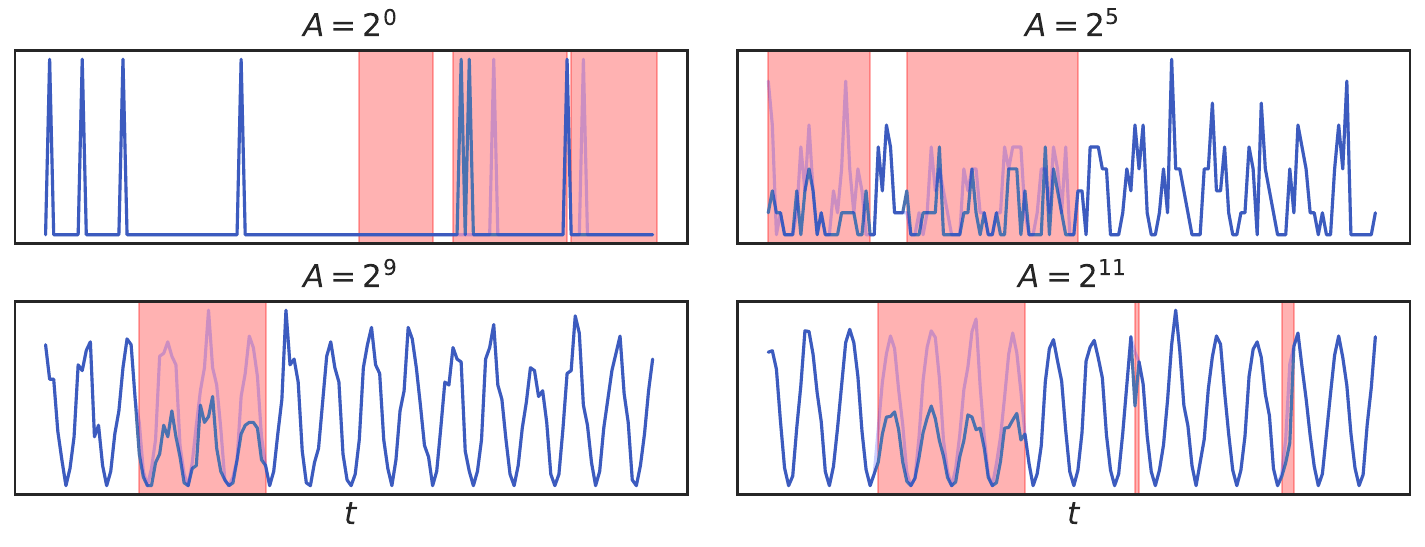}}
  \vspace{-0.4cm}
  \caption{Example time series simulated using our generative process, where the amplitude $A$ controls the count level.
  Injected anomalies with reduction rate $r=0.5$ are highlighted in red.}
\label{fig:synthetic_data}
\end{figure}

To construct the dataset used in our experiments, we set $f=1$ and define a transition matrix with $T_{00}=0.995$, $T_{11}=0.95$, and $\Delta t=0.1$ throughout.
We generate datasets with pairwise combinations of two variables: $A \in \{2^{-2},\dots, 2^{12}\}$ and $r \in \{0.1, 0.5, 0.9, 1.0 \}$. 
This results in a total of 60 time series covering a wide range of count levels and anomaly severities.
We account for potential variance in this set-up by sampling five independent sets of anomaly locations in each time series.
Sample segments from the generated time series are illustrated in Figure \ref{fig:synthetic_data}.

\section{Benchmark Set-up}\label{sec:experimental_setup}

Equipped with the data-generating procedure described above, the second main contribution of our work is an extensive evaluation of how widely-used AD techniques perform on datasets containing time series with varying count levels.

\subsection{Categorisation of AD Methods}\label{sec:ad_methods}

We start by briefly introducing the classes of AD techniques that will feature in our evaluation.
These were selected on the basis of being widely-used by practitioners in multiple domains, and follow the categorisation suggested in \cite{schmidlAnomalyDetectionTime2022}.

\noindent \textbf{Forecasting Methods:}
Assuming the use of a probabilistic forecasting technique, methods in this category generate a predictive distribution $p_t(\cdot)$ for each future time step, which is then compared to the incoming actual data $x_t$ via a scoring function that emits an anomaly score.
Widely-used scoring functions include:
\begin{itemize}
    \setlength\itemsep{-0.2em}
    \item \textbf{Scaled Absolute Forecasting Error.}
    This is computed as $|x_t - \mu_t|/\sigma_t$, where $\mu_t$ and $\sigma_t$ are the mean and standard deviation of the predictive distribution.
    \item \textbf{Quantile Score.} This is calculated by $\texttt{max}((|q_{50} - x_t|-(q_{50} - q_l))/(q_{50} - q_l), 0)$, where $q_{50}$ is the median and $q_l$ is the $l$-th quantile (e.g. $l=5$) of the  predictive distribution.
    \item \textbf{Negative Residual Score.} This is calculated as $-(x_t - \mu_t)/\sigma_t$, and is responsive only to signal drops. 
\end{itemize}

\noindent \textbf{Distance Methods:}
These methods utilize the distances between segments of a time series to identify anomalies.
For example, the $k$-NN algorithm computes pairwise distances between the target segment and preceding segments \cite{angiulliFastOutlierDetection2002}, and the resulting distance can be treated as an anomaly score.
The Matrix Profile algorithm~\cite{yehMatrixProfileAll2016} works similarly to $k$-NN, except that distances are standardized and the computations are highly optimized with $k$ fixed at 1. 

\noindent \textbf{Reconstruction Methods:}
Algorithms such as autoencoders and variational autoencoders~\cite{kingmaAutoEncodingVariationalBayes2013} learn a model that compresses segments of the original time series into a lower-dimensional latent space, and decompresses them back with minimized losses.
Assuming it would be easier for the model to reconstruct the target segment if it were normal, the reconstruction loss naturally translates to an anomaly score.

\noindent \textbf{Distribution Methods:}
Distribution-based methods operate under the assumption that anomalies are typically located in the tails of the data distribution.
Two popular approaches falling under this category are the Histogram-Based Outlier Score (HBOS)~\cite{goldsteinHistogrambasedOutlierScore2012} and Empirical-Cumulative-distribution-based Outlier Detection (ECOD)~\cite{liECODUnsupervisedOutlier2022}.

\noindent \textbf{Tree Methods:}
Finally, methods such as Isolation Forest \cite{liuIsolationForest2008} determine the number of random feature splits needed to isolate a sample from all others.
Hence, the number of required splits serves as an implicit anomaly score.

\subsection{Models under evaluation}\label{sec:algorithms}
Without loss of generality, we select a representative algorithm from each category to provide a general overview of how each class of methods adapts to low-count data.
Specifically, the algorithms used are: (i) LSTM-Forecasting with three common scoring functions (\textsc{absolute-error}, \textsc{quantile}, \textsc{negative-residual}), (ii) \textsc{matrix-profile}, (iii) \textsc{auto-encoder}, (iv) \textsc{ecod}, and (v) \textsc{isolation-forest}.

To further contextualise the performance of each algorithm in relation to an optimal reference, we also include results for a Bayes optimal classifier (\textsc{boc}) \cite{norvigArtificialIntelligenceModern}.
This model has access to complete knowledge of how the data and anomalies are generated (this is a key benefit of constructing our own synthetic data, as detailed in Section~\ref{sec:synthetic_data}), and serves as an upper bound on expected performance in the challenging low-count contexts.
We also include a simple baseline, \textsc{zero-run-length}, which only indicates an anomaly when a tally of sequential zero values exceeds a threshold.

\subsection{Evaluation Metrics}

We evaluate performance with the Area Under The Precision-Recall Curve (AUPRC) and $F_{1}$ metrics since these are more resilient to imbalanced classes (AD will have far fewer abnormal data points than normal data points). 
We note that similar conclusions are gleaned from our analysis of alternative measurements, such as range-based precision, recall, and their extensions, but we do not include these in our evaluation due to space limitations. 
\section{Results and Discussion}
\label{sec:results}
In this section, we present a benchmark evaluation of how different TSAD algorithms perform on our synthetic dataset.
We also investigate why certain methods perform better than others on low-count data, and offer insights that could guide practitioners towards deploying more robust models.
This analysis leads to our recommendation of applying smoothing operators to raw anomaly scores, which can dramatically improve performance in low-count settings.
We validate the utility of our findings in a practical setting via an experiment on the M5 competition dataset describing Walmart retail data.

\subsection{Model Comparison on Synthetic Data}
\begin{figure}[t!]
  \centering
  \centerline{\includegraphics[width=8.5cm]{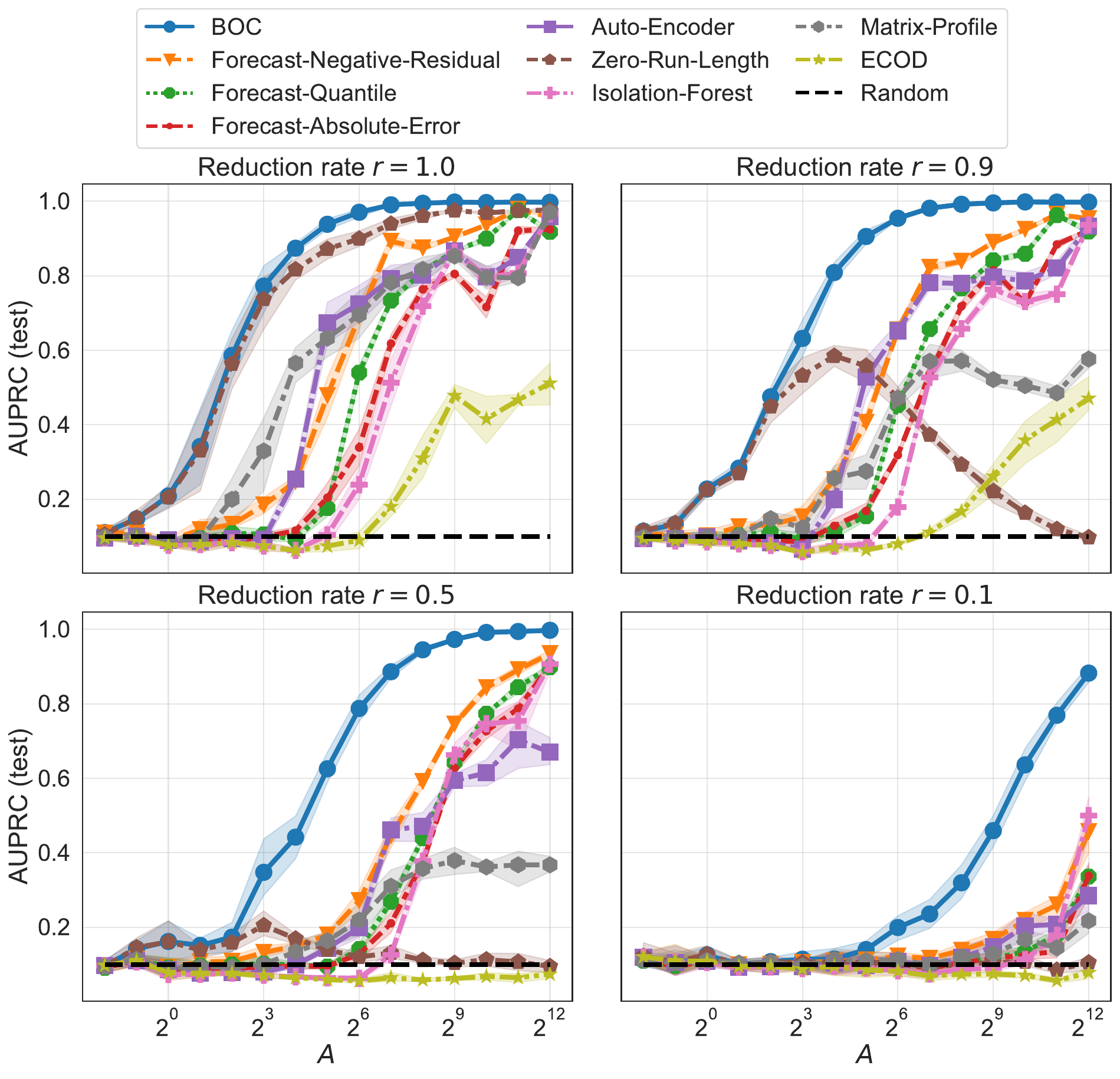}}
 \vspace{-0.4cm}
 \caption{AUPRC for tested methods across count levels.}
\label{fig:result_all_low_volume}
\end{figure}

Figure \ref{fig:result_all_low_volume} shows the AUPRC for all tested TSAD algorithms under different count levels.
The error bars indicate the variance derived from running each algorithm on five instantiations of the synthetic dataset (see Section~\ref{sec:exp_dataset}).
As highlighted earlier, \textsc{boc} denotes an upper bound on the performance that can be achieved for each anomaly severity level. 

The results indicate that the algorithms perform worse as the count level decreases, with some algorithms performing no better than random ($\textrm{AUPRC}=0.1$) when $A < 2^{4}$.
We believe this deficiency has not been clearly acknowledged in other research works, and the gap may easily be obscured by a model's superior performance on higher-count time series for general datasets.
The rankings of individual algorithms seem mostly consistent across different anomaly severities;
in the remainder of this subsection, we provide additional insight that explains their behaviour.

Forecasting methods achieve the best performance across all anomaly severity levels in the highest-count time series.
This matches the findings presented in \cite{schmidlAnomalyDetectionTime2022}.
However, their performance degrades very quickly for low-count time series, approaching random performance at count levels below $2^{6}$.
This behaviour can be explained by Figure \ref{fig:forecasting_failed} \textemdash{} while the deviation induced by the second anomaly can be easily detected, the predicted confidence interval during the first anomaly includes 0, yielding low anomaly scores in this region.

\textsc{matrix-profile} (and distance models generally) develop a `memory' that grows with the time series, which can be problematic when multiple anomalies are expected.
This is because the first expression of an anomaly will tend to produce a large anomaly score (since it is not yet contained in the history), while subsequent expressions will get smaller scores owing to the existence of a similar signature in the data.
This would be expected in the top examples in Figure~\ref{fig:synthetic_data} where the highlighted anomalies resemble one another.
To mitigate this issue, we update its scoring function to only consider anomaly-free areas in the history.
Nonetheless, we still observe poor performance in low-count areas where both anomalous and normal data may contain long runs of intermittent or zero values.

\textsc{auto-encoder} models work well for severe anomaly types, and only lag behind forecasting for the highest count levels.
However, performance tails off once again for lower count levels.
Our hypothesis for why this happens follows a similar reasoning to \textsc{matrix-profile} - anomalies would look similar to normal areas for low-count time series, and a reconstruction model trained on normal areas may just as effectively reconstruct anomalous areas.
Similarly, \textsc{isolation-forest} works nicely for high-count data, but the summary statistics from normal segments may once again look very similar to anomalous ones for low-count areas.

\textsc{ecod} performs most poorly among all tested algorithms.
From our investigations, we deduced that distribution-based methods are not suitable for seasonal time series because more samples concentrated around the peaks and troughs can lead to heavy probabilities around the low-count areas.
When an anomaly occurs in the vicinity of a trough, it may still fall within the expected range for the data. Pre-processing would be required in order to reduce the impact of seasonality.
Finally, the trivial \textsc{zero-run-length} baseline naturally performs well for the $r=1.0$ anomalies it is engineered to detect, but does not generalise.
This confirms that simple baselines may only be suitable if anomaly behaviours are known in advance.

\subsection{Anomaly Score Smoothing}\label{sec:smoothing}

\begin{figure}[t!]
  \centering
  \centerline{\includegraphics[width=0.95\linewidth]{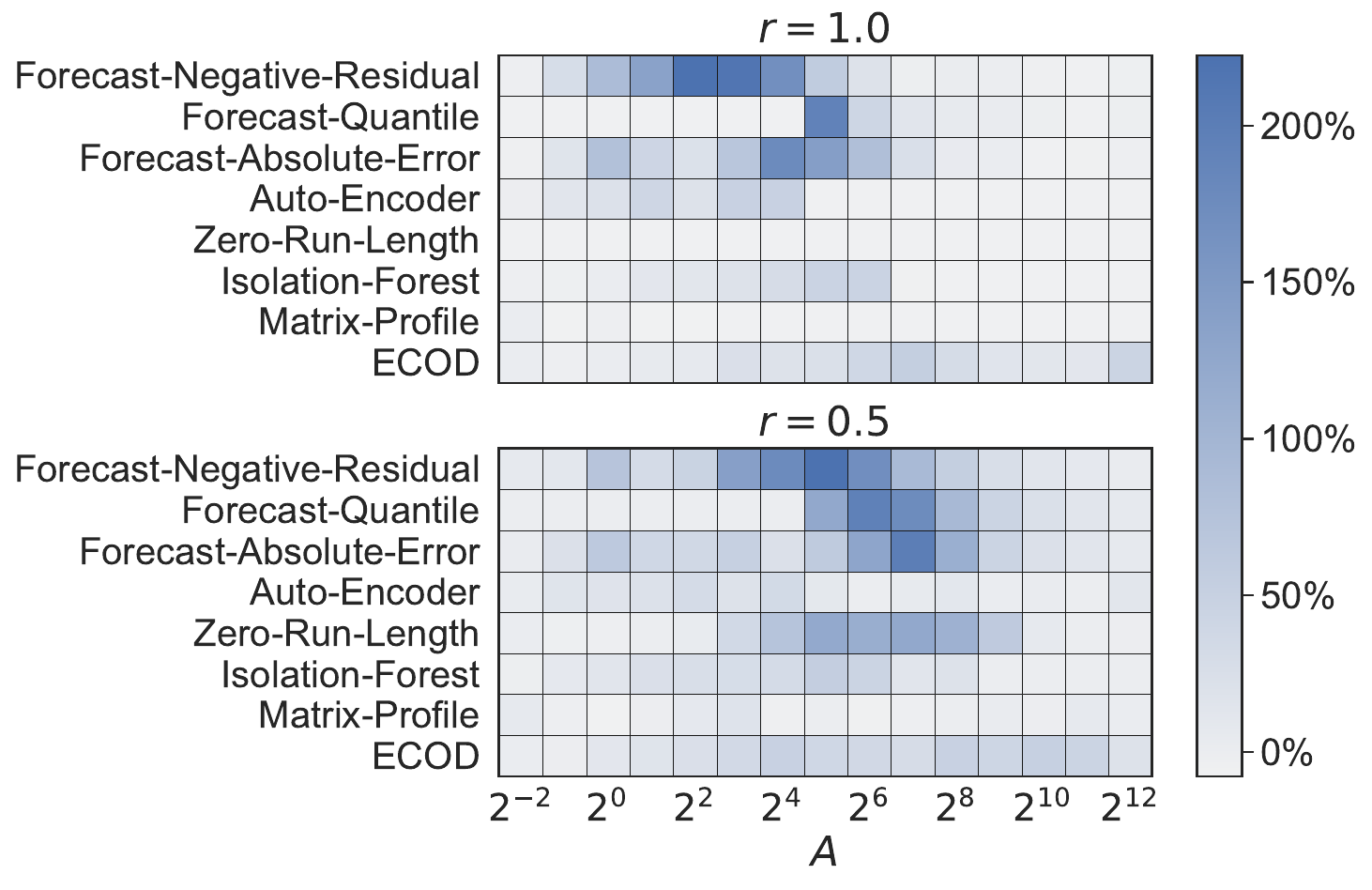}}
 \vspace{-0.4cm}
 \caption{Rate of improvement on AUPRC achieved by applying EMA smoothing to the raw anomaly scores.}
\label{fig:agg_result_all_low_volume}
\end{figure}

Another aspect that contributes to poor performance is the greater noise exhibited in the obtained anomaly scores.
This corresponds to the lower signal-to-noise ratio observed when the count is low.

In this light, we experiment with a standard exponential moving average (with decay rate 0.125, based on $T_{11}$) applied on top of the raw anomaly scores in order to smooth them.
In Figure~\ref{fig:agg_result_all_low_volume}, for each combination of AD method and count level, we illustrate the improvement on AUPRC obtained by applying EMA smoothing to the raw anomaly scores.
This is computed as a percentage improvement over the original results (Figure~\ref{fig:result_all_low_volume}) without any smoothing.
The results indicate that anomaly score smoothing improves AUPRC in most settings, and crucially never harms performance.
The performance improvements are particularly notable for \textsc{forecast-negative-residual} in low-count settings.

There is however an implied risk that smoothing functions introduce a lagging effect that could negatively impact time-to-detection in online settings.
Time-to-detection (TTD) measures the delay in detecting an anomaly, and is particularly crucial in applications where immediate remediation or diagnosis is required.
To assess this risk, we set up another experiment to visualise the trade-offs between performance and TTD that arise from smoothing.
For ease of exposition, we take the winning model \textsc{forecast-negative-residual} from the previous comparison and show how the trade-off between $F_{1}$ score and TTD changes for different smoothing operators.
We experiment with exponential moving average (EMA) and a variety of functions applied to sliding windows: mean (SWAvg), median (SWMed), maximum (SWMax) and minimum (SWMin).

\begin{figure}[t!]
  \centering
  \centerline{\includegraphics[width=8cm]{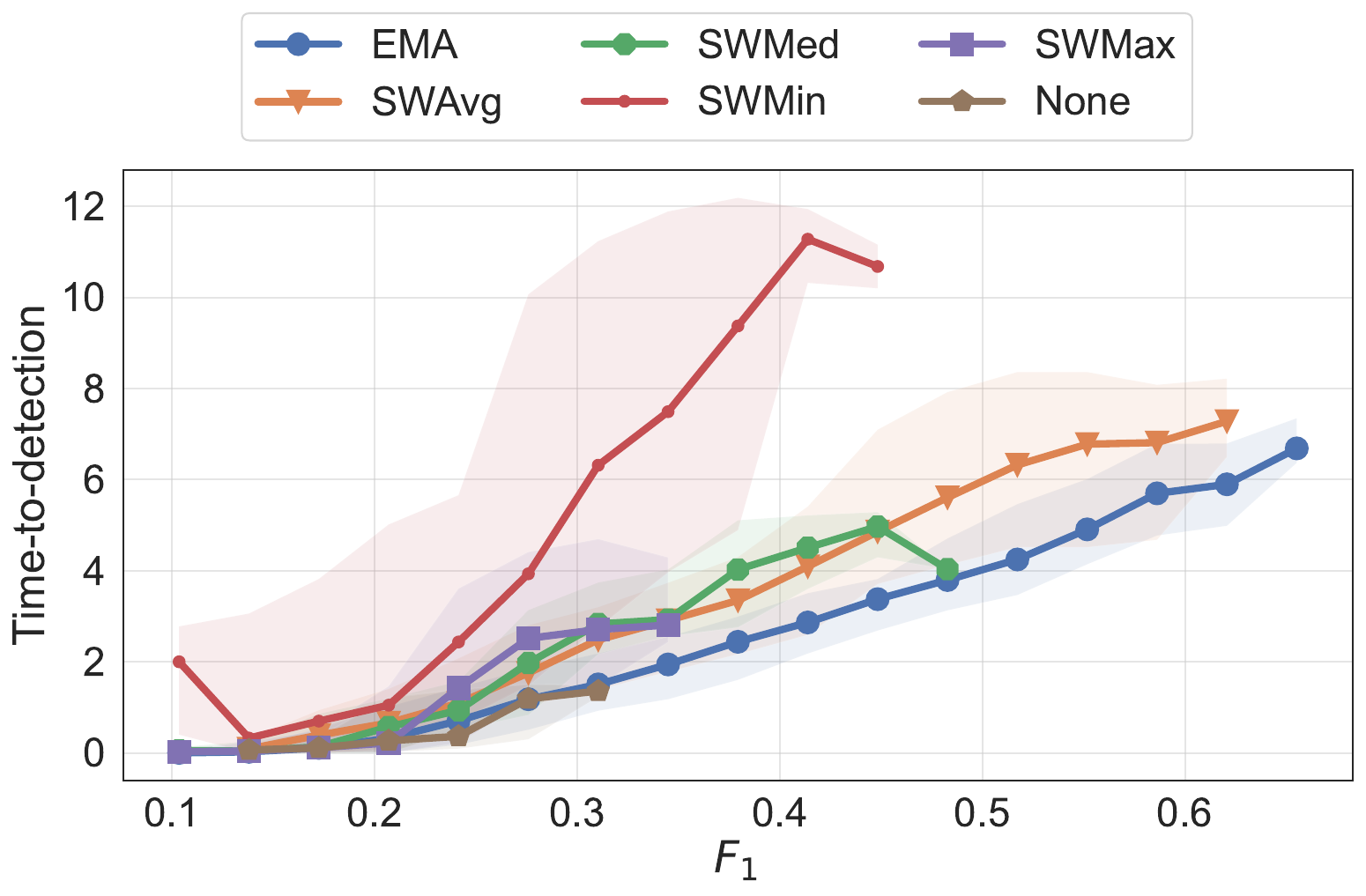}}
 \vspace{-0.4cm}
\caption{$F_{1}$ \textit{vs.} TTD for different smoothing functions.}
\label{fig:time_to-detection}
\end{figure}

The results are plotted in Figure~\ref{fig:time_to-detection}, where each data point represents the results obtained for a specific choice of threshold.
The synthetic data is generated with parameters $A=2^5$ and $r=0.5$, which simulate a medium anomaly type in a common low-count time series. 
The results indicate that EMA maintains the best trade-off between $F_{1}$ and TTD for all threshold settings.
Note that the best TTD is initially achieved when no smoothing is applied, but at the expense of very low accuracy. 
This is especially visible in the low-count settings where the performance peaks at $F_{1}$ score of 0.3 for any choice of threshold.
In light of these results, we recommend anomaly score smoothing with EMA as a means to improve performance on low-count time series.

An alternate approach is to aggregate the data at larger temporal granularity upfront, and monitor the re-sampled data streams.
However, the penalty on TTD is expected to be greater using this approach, and supporting multiple temporal aggregations may require changes to model training pipelines.
Additionally, by leveraging signal processing on model scores, the developed solution does not require fundamentally new technologies for training or deployment, and can be trivially deployed to increase reliability and scope of anomaly detection systems in production environments. 

\subsection{Application to M5 Walmart dataset}
So far, we have benchmarked the performance of different TSAD methods using our synthetic dataset, leading to our recommendation of smoothing raw anomaly scores. 
In this section we carry out a similar experiment on the M5 forecasting competition data~\cite{Makridakis2022Comp}, which consists of 1913 days of sales data for a selection of Walmart items.

\begin{figure}[t!]
  \centering
  \includegraphics[width=8.5cm]{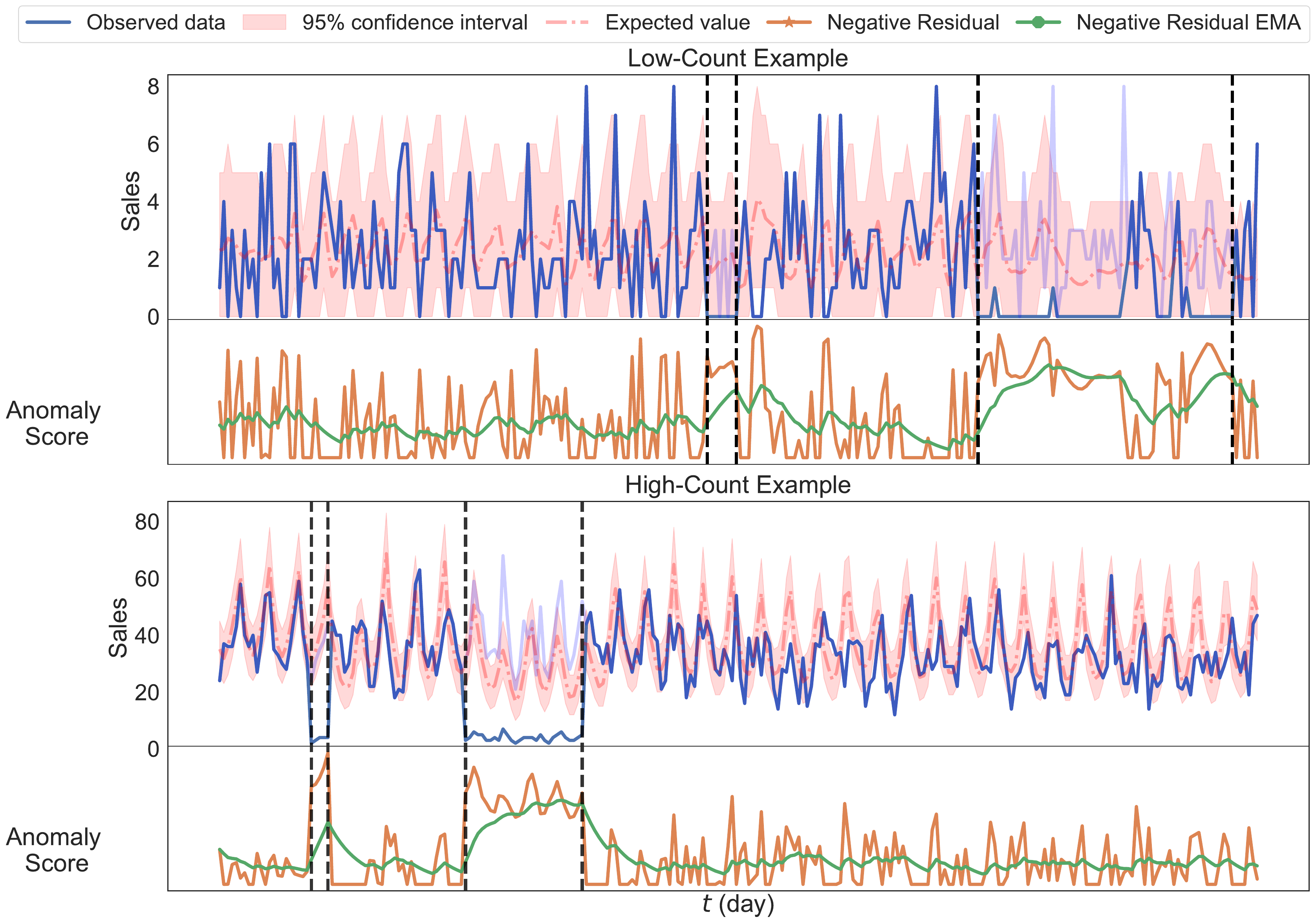}
  \vspace{-0.4cm}
\caption{
Example of injected anomalies in the M5 dataset.
}  
\label{fig:m5_data}
\end{figure}

We sampled 55 time series from this dataset covering a wide spectrum of count levels. Given that we no longer have full control on the count levels, we grouped the time series into four bins of count levels to allow us to stratify and better visualise the results.
The dataset originates from forecasting research and lacks labeled anomalies, so we adopt the approach detailed in Section \ref{sec:anomaly_injection} to inject artificial anomalies (see Figure \ref{fig:m5_data}).
These represent plausible scenarios where sales can abruptly drop. For the low-count time series, the smoothing stabilises the noisy anomaly scores that would otherwise hinder the use of thresholds for distinguishing between normal and anomalous data.
On the other hand, for high-count data smoothing introduces a lagged effect to the anomaly scores beyond the actual anomaly area, that could be misinterpreted as false positives when using performance metrics with no temporal tolerance.

%

\begin{figure}[t!]
  \centering
  \includegraphics[width=8.5cm]{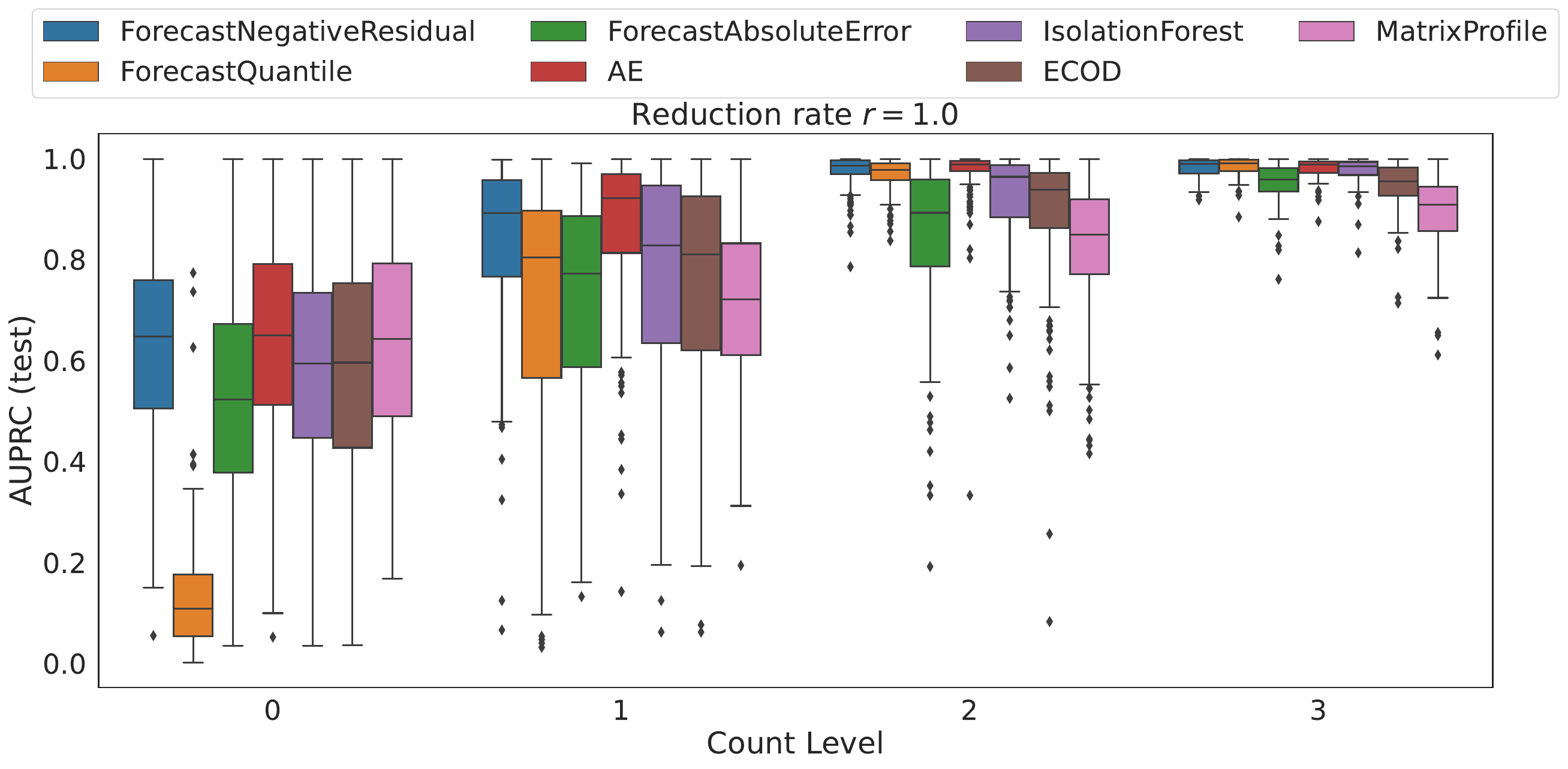}
  \vspace{-0.4cm}
\caption{AUPRC on M5 dataset at varying count levels.}
\label{fig:m5_result}
\end{figure}

The results are plotted in Figure \ref{fig:m5_result}.
Given that the time series in the sub-sampled M5 dataset are much shorter than our synthetic data, repeated experiments with different anomaly injections lead to higher variance in the results. Nevertheless, we can still clearly observe the same degradation in performance when algorithms are applied to low-count time series. 
The results additionally show that some algorithms are more sensitive to count levels than others.
For example, \textsc{forecast-quantile} works poorly with the lowest count level, but competitively with higher count cases.
This is likely because the predicted median and lower quantile become indistinguishable in low-count scenarios.
Although not shown, we observe similar results for other settings of $r$.
As before, \textsc{forecast-negative-residual} and \textsc{auto-encoder} emerge as top performers, and with the aid of anomaly score smoothing perform reasonably well even in low-count contexts where general solutions otherwise fail. 

\section{Conclusion}
\label{sec:conclusion}

Popular TSAD models can miss trivial-looking anomalies in low-count contexts, and this work delivers the first comprehensive investigation surrounding the causes of this low performance ceiling.  
To enable this, we developed a new benchmarking system that allows for broad exploration and exact evaluation in low-count TSAD.
This played a critical part in identifying key failure points in various models, which could not otherwise be captured from standard datasets due to labelling noise~\cite{Keogh2021} (and is further exacerbated in low-count contexts due to poor signal-to-noise ratios).   

This analysis also led us to recommend an intuitive mitigation, whereby we demonstrated that employing a smoothing function on top of generated anomaly scores can consistently improve performance irrespective of model class.
We trust that our findings will not only serve as a guide for practitioners monitoring similarly diverse collections of time series, but also instil greater research interest in verifying the robustness of AD techniques under challenging settings. 

{
\bibliographystyle{IEEEbib}
\newlength{\bibitemsep}\setlength{\bibitemsep}{.2\baselineskip plus .05\baselineskip minus .05\baselineskip}
\newlength{\bibparskip}\setlength{\bibparskip}{0pt}
\let\oldthebibliography\thebibliography
\renewcommand\thebibliography[1]{%
  \oldthebibliography{#1}%
  \setlength{\parskip}{\bibitemsep}%
  \setlength{\itemsep}{\bibparskip}%
}
\small
\bibliography{references_no_detail} 

}


\end{document}